\documentclass[letterpaper, 10 pt, conference]{ieeeconf}  
\usepackage{graphicx}
\usepackage{tabularx}

\IEEEoverridecommandlockouts                              

\usepackage{graphicx}
\usepackage{float}
\usepackage{amsmath}
\usepackage{algorithmic}
\usepackage{amsfonts}
\usepackage{times}
\usepackage{algorithm}
\usepackage{xcolor} 
\usepackage[normalem]{ulem}
\usepackage{comment}
\newcommand{\raj}[1] {{\color{blue} #1} }
\newcommand{\spr}[1] {{\color{red} #1} }

\usepackage{multirow}
\title{\LARGE \bf
Stereo Visual Odometry with Deep Learning-Based Point and Line Feature Matching using an Attention Graph Neural Network
}

\author{Shenbagaraj Kannapiran$^{\dagger 1}$, Nalin Bendapudi$^{2}$, Ming-Yuan Yu$^{\ddagger 3}$, Devarth Parikh$^{2}$,\\ Spring Berman$^1$, Ankit Vora$^{2}$, and Gaurav Pandey$^{2}$
\thanks{$^{\dagger}$Work done during an internship at Ford Next LLC.}
\thanks{$^{\ddagger}$Work done while employed 
at Ford Next LLC.}
\thanks{$^{1}$Shenbagaraj Kannapiran and Spring Berman are with the School for Engineering of Matter, Transport and Energy, Arizona State University, Tempe, AZ 85287, USA
        {\tt\small shenbagaraj@asu.edu, spring.berman@asu.edu}}%
\thanks{$^{2}$Nalin Bendapudi, Devarth Parikh, Ankit Vora and Gaurav Pandey are with 
Ford Next LLC, Ann Arbor, MI 48109, USA
        {\tt\small nbendapu@ford.com, dparikh9@ford.com, avora3@ford.com, gpandey2@ford.com}}%
\thanks{$^{3}$Ming-Yuan Yu is with Qualcomm, Novi, MI 48377, USA
        {\tt\small myyu@umich.edu}}%
}

\begin{document}

\maketitle
\thispagestyle{empty}
\pagestyle{empty}

\begin{abstract}

Robust feature matching forms the backbone for most Visual Simultaneous Localization and Mapping (vSLAM), visual odometry, 3D reconstruction, and Structure from Motion (SfM) algorithms. However, recovering feature matches from texture-poor scenes is a major 
challenge and still remains an open area of research. In this paper, we present a Stereo Visual Odometry (StereoVO) technique based on point and line features which uses a novel feature-matching mechanism based on an Attention Graph Neural Network that is designed to perform well even under adverse weather conditions such as fog, haze, rain, and snow, 
and dynamic lighting conditions such as  nighttime illumination and glare scenarios. 
We perform experiments on multiple real and synthetic datasets to validate  our  method's ability to perform StereoVO under low-visibility weather and lighting conditions through robust point and line matches. The results demonstrate that our method achieves more 
line feature matches than state-of-the-art line-matching algorithms, which when complemented with point feature matches perform consistently well in adverse weather and dynamic lighting conditions. 
\end{abstract}

\section{INTRODUCTION}


With the advances in development and deployment of 
self-driving vehicles and mobile robots, there is a growing need for high-resolution, 
accurate visual odometry algorithms that can be deployed on low-cost camera sensors. 
Although 
existing localization algorithms perform well under ideal conditions, they usually tend to fail or under-perform in adverse weather conditions such as fog, rain, and snow and in dynamic lighting conditions such as glare and nighttime illumination. The development of visual odometry algorithms that perform effectively under such conditions remains an open area of research. 

\begin{figure}[t]
  \centering
  \includegraphics[width=\linewidth]{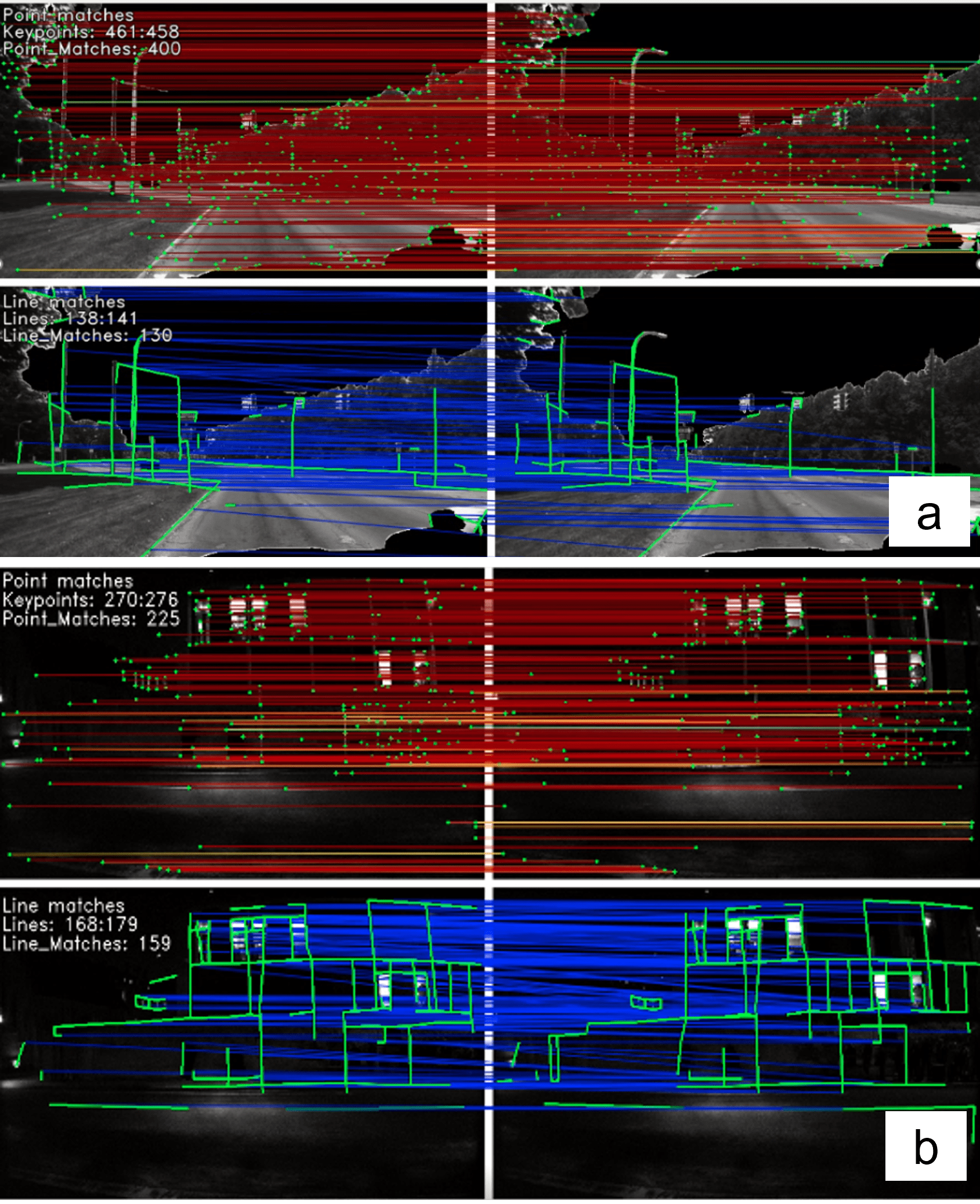}
  \caption{
  Grayscale images 
   of 
  (a) a clear-sky scenario from the Ford AV dataset~\cite{agarwal2020ford}; (b) a nighttime scenario from the Oxford Car dataset~\cite{RobotCarDatasetIJRR}. Rows 1 and 3 show 
  point matches between frames $i$ and $i+1$; rows 2 and 4 show 
  line matches between frames $i$ and $i+1$. Point and line matches were generated by the method presented in this paper (Method 2 in Section \ref{sec:exp}).}
  \label{fig:teaser}
\end{figure}

Existing vision-based localization algorithms rely primarily on conventional point features, such as SIFT~\cite{c1}, SURF~\cite{c2}, and ORB~\cite{c3}, or learning-based point features, such as SuperPoint~\cite{c4} and LIFT~\cite{c5}, to perform temporal feature matching or learn alternate representations using monocular camera(s)~\cite{Hausler2022, voodarla2021} 
and obtain camera pose estimates.
However, point features become unreliable in the adverse conditions mentioned above. To overcome this, we develop a visual odometry technique that includes line segment features in addition to 
point features (see example application in Fig. \ref{fig:teaser}). In foggy scenes where conventional point features like ORB and SIFT are unable to detect enough features, learning-based feature detectors like SuperPoint perform well. Line feature detectors such as SOLD2 \cite{pautrat2021sold2} and L2D2 \cite{abdellali2021l2d2} also perform well in such scenarios.  
However, most point feature detectors, including ORB, SIFT, and SuperPoint, tend to demonstrate poor performance in nighttime scenarios. In comparison, the 
line features detected by state-of-the-art line detectors such as SOLD2 and L2D2 tend to remain consistent in scenes with low illumination, 
thereby indicating the need to leverage line features for nighttime conditions. 

Given the importance of line features, the next step in integrating line features with the visual odometry framework is to perform feature-matching for line features. Existing techniques either utilize line-based descriptors or apply point descriptors to 
points sampled from detected lines. Both types of techniques rely on visual descriptors, which tend to fail in texture-poor scenes, justifying the need for a line-matching solution that is constrained by the positions of line features and visual cues. Position-constrained line-matching ensures the prevention of line feature mismatches, particularly in scenes where point features are sparse or similar structures appear repeatedly, such as trusses of a bridge and windows in urban high-rises.  

Apart from vision-based sensors, inertial sensors such  as Inertial Measurement Units (IMUs) can be used to aid the system to perform Visual-Inertial Odometry (VIO), which yields better accuracy. However, we restricted our focus to just vision-based systems to showcase our method's capabilities without the aid of other such sensors. We developed our method with the goal of easily integrating it into a standard self-driving research vehicle for real-time deployment, and hence we assume that our visual-odometry based pose estimate will ultimately be fused with GNSS (Global Navigation Satellite System) and IMU-based pose estimates in an extended Kalman filter-based framework to provide more accurate pose estimates.

Our contributions can be summarized as follows:
 \begin{itemize}
    \item We developed a novel line-matching technique using an Attention Graph Neural Network that is capable of acquiring robust line matches in feature-poor scenarios by sampling and detecting self-supervised learning-based point features along the lines with encoded position constraints.  
    \item We integrated point features and fine-tuned line features in a Stereo Visual Odometry framework to maintain consistent performance in adverse weather and dynamic lighting conditions and compared the performance of our method to that of 
    state-of-the-art point and line feature matching techniques.
    
\end{itemize}

We discuss 
related work in Section II, give 
a technical description of our approach in Section III, and describe experiments and results in Section IV. Section V concludes the paper and provides an outlook on future work.


\section{RELATED WORK}

In this section, we give an overview of previous work related to visual odometry, graph matching, point feature matching, and line matching. Given the amount of prior research, particularly in the visual odometry field (e.g., 
\cite{c7,c8,c9}), a compilation of all 
existing visual odometry algorithms is beyond the scope of this paper. Visual odometry (VO) techniques are usually classified as either direct VO or feature-based VO. Feature-based solutions are primarily used for their reliability, high accuracy, and robustness, and will therefore be the focus of this paper.

\subsection{Point feature detection and matching}

Point feature detection lies at the heart of most vision-based algorithms. The paper~\cite{c10} presents a comprehensive survey of different classical feature detectors such as SIFT, ORB, and SURF and learning-based detectors such as LF-Net~\cite{c11} and SuperPoint and compares their performance on three evaluation tasks in terms of robustness, repeatability, and accuracy. Classical feature-matching techniques usually involve finding descriptors, 
matching them using a nearest neighbor search, and finally removing outliers to obtain robust matches. Over time, researchers moved towards developing more robust and accurate feature detectors and descriptors to improve matching. Then 
graph neural network (GNN)-based matching systems such as SuperGlue \cite{c6} were developed, which outperformed all existing feature matching techniques by using an attentional GNN. The attention mechanism enables the GNN to selectively focus on the most relevant features (i.e., nodes and edges) when comparing two graphs, which improves the accuracy of feature matching and its robustness to noisy or incomplete graphs. 
For this reason, we used SuperGlue for point feature matching in our Stereo Visual Odometry framework.

\subsection{Line feature detection and matching}

Line feature detection and matching is a well-researched topic. Classical line detector algorithms rely on geometric constraints to extract lines and find correspondences. Similarly, line segment descriptors can be constructed from the appearance of a neighborhood of the detected line, without resorting to any other photometric or geometric constraints such as the mean--standard deviation line descriptor (MLSD) \cite{wang2009msld}, which constructs the line descriptors by computing the mean and variance of the gradients of pixels in the neighboring region of a line segment. The work \cite{zhang2013efficient} proposes a 
Line Band Descriptor (LBD) that computes gradient histograms over bands for improved robustness and efficiency. Recent advancements in learning-based line segment descriptors, e.g., LLD \cite{vakhitov2019learnable} and DLD \cite{lange2019dld}, demonstrate excellent performance with the use of a convolutional neural network (CNN) to learn the line descriptors. In 
\cite{abdellali2021l2d2}, the authors propose a novel line segment detector and descriptor, Learnable Line Detector and Descriptor (L2D2), which enables 
efficient extraction and matching of 2D lines via the angular distance of 128-dimensional unit descriptor vectors. The paper 
\cite{ma2020robust} presents a novel Graph Convolutional Network-based line segment matching technique 
that learns local line segment descriptors through end-to-end training. 

In \cite{pautrat2021sold2}, the authors propose SOLD2, a self-supervised learning-based line detector that is similar to SuperPoint and does not require any annotation, enabling the system to generalize to multiple scenarios. 
For this reason, we chose SOLD2's line detector module as a baseline for our method. SOLD2 also includes a line-matching algorithm to enable occlusion awareness. However, unlike SuperGlue, SOLD2's matching algorithm does not take advantage of the position information of the features, which is critical in scenes that contain 
repetitive structures such as 
windows in urban high-rises.



\subsection{Visual SLAM / Odometry with 
point and line features}

As described in \cite{zuo2017robust,yang2019visual}, visual SLAM methods that incorporate both point and line features  have been developed to improve localization accuracy and computational efficiency over conventional point-based approaches in challenging scenarios, making the VO pipeline more comprehensive and robust to real-world conditions. 
One example is the visual-inertial SLAM method 
in \cite{xu2022eplf}, which includes several enhancements in line detection and an optical flow-based line feature tracker.  
Another is the line classification methodology for a Visual Inertial Odometry system that is presented in \cite{xu2022leveraging}, which exploits the distinctive characteristics of structural (parallel) and non-structural (non-parallel) line features to develop a two-parameter line feature representation, leading to more efficient SLAM computations. However, despite the benefits afforded by using both point and line features, these visual SLAM techniques often exhibit poor performance in scenarios with repeated, similar-looking point and line features, such as those found in traffic environments (e.g., building facades, pedestrian crosswalks). The design of our StereoVO technique was motivated in part by this limitation.

\section{StereoVO WITH POINTS AND LINES}

Our proposed 
StereoVO framework is developed to perform well in texture-poor scenarios and relies on tracking a set of point and line correspondences. 
The framework is based on the SuperGlue~\cite{c6} network, with an additional constraint (constraint (3) in Section \ref{sec:NotationDef}) that greatly improves performance. 
 We first provide a 
overview of the StereoVO framework, followed by the notation and definitions that we use in our Attention Graph Neural Network architecture. This is followed by a description of the Optimal Matching layer for both point and line features, and a brief summary of how to obtain 
pose estimates from the point and line correspondences.

\begin{figure}[]
  \centering
 \includegraphics[width=0.9\linewidth]{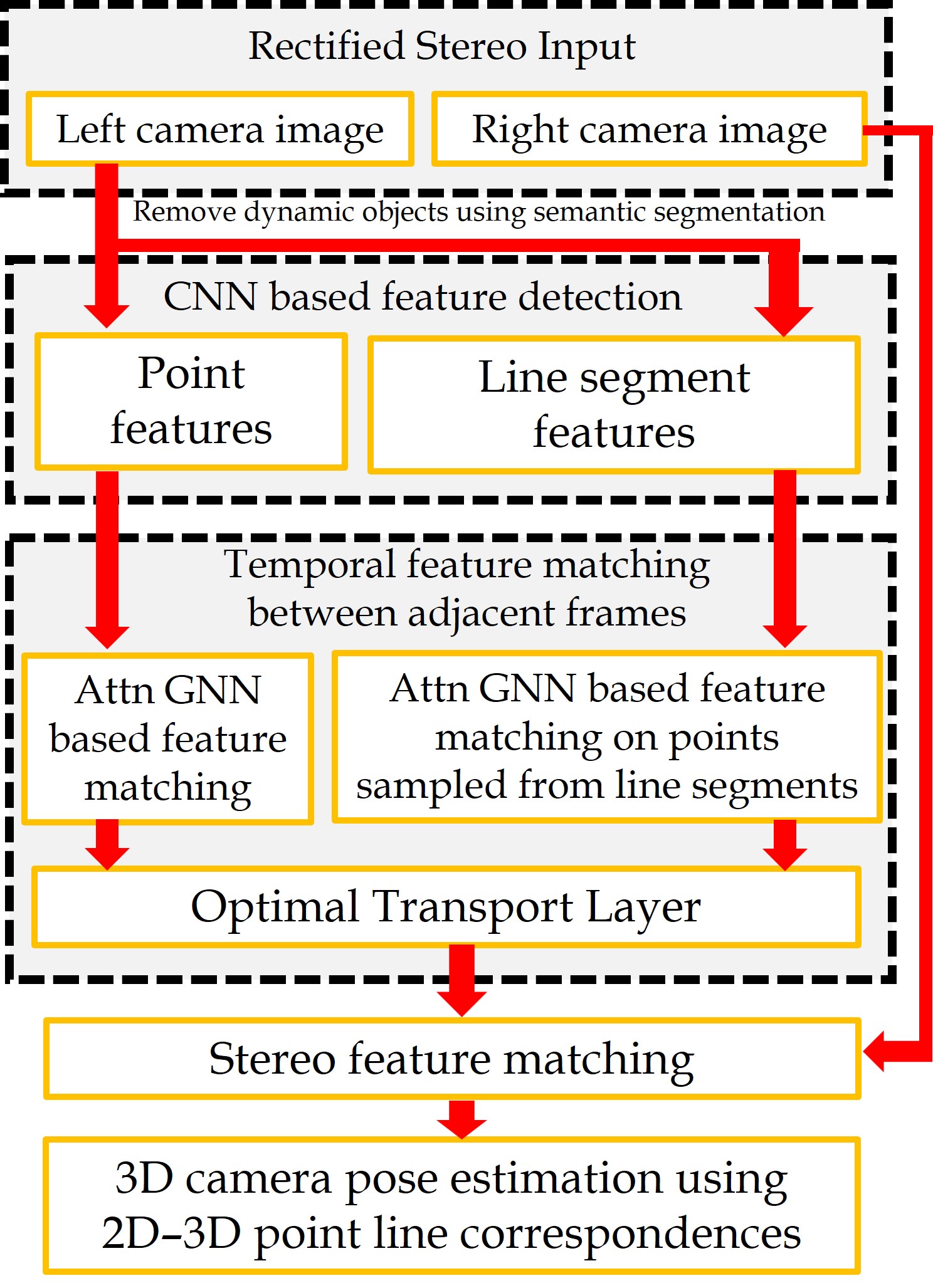}
  \caption{Outline of the proposed Stereo Visual Odometry framework.}
  \label{fig:outline}
\end{figure}

\subsection{Overview}

An outline of the proposed StereoVO framework is shown in Fig. \ref{fig:outline}. The stereo images obtained from the camera are initially undistorted. The left camera image is 
used to obtain point and line matches, and the right camera image is used to obtain 3D points and 3D line estimates from disparity maps generated using the stereo images. 

To improve the accuracy of 
StereoVO, the left camera image is pre-processed using a semantic segmentation algorithm to remove dynamic objects, such as cars and pedestrians, thereby generating a mask that highlights static features in the scene.
In StereoVO, focusing on stable features 
improves the reliability, precision, and robustness 
of 
the camera pose estimates. 
We employ SegFormer \cite{xie2021segformer}, a state-of-the-art semantic segmentation algorithm, out of the box to mask the classes of interest.


In the next step, we perform point and line segment feature detection on the masked input images. Since our goal is to implement the framework on a full-size autonomous vehicle and ensure that it is capable of performing well under adverse weather and dynamic lighting conditions, we tested 
a 
variety of 
point feature detectors, including SIFT, ORB, and LIFT, in example scenarios with such conditions
and chose 
SuperPoint since it outperformed the others. 
SuperPoint is a CNN-based self-supervised framework that is trained on the MS-COCO dataset \cite{lin2014microsoft} and performs real-time point feature detection out of the box without any fine-tuning. 

\begin{figure*}[t]
  \centering
  \includegraphics[width=\textwidth]{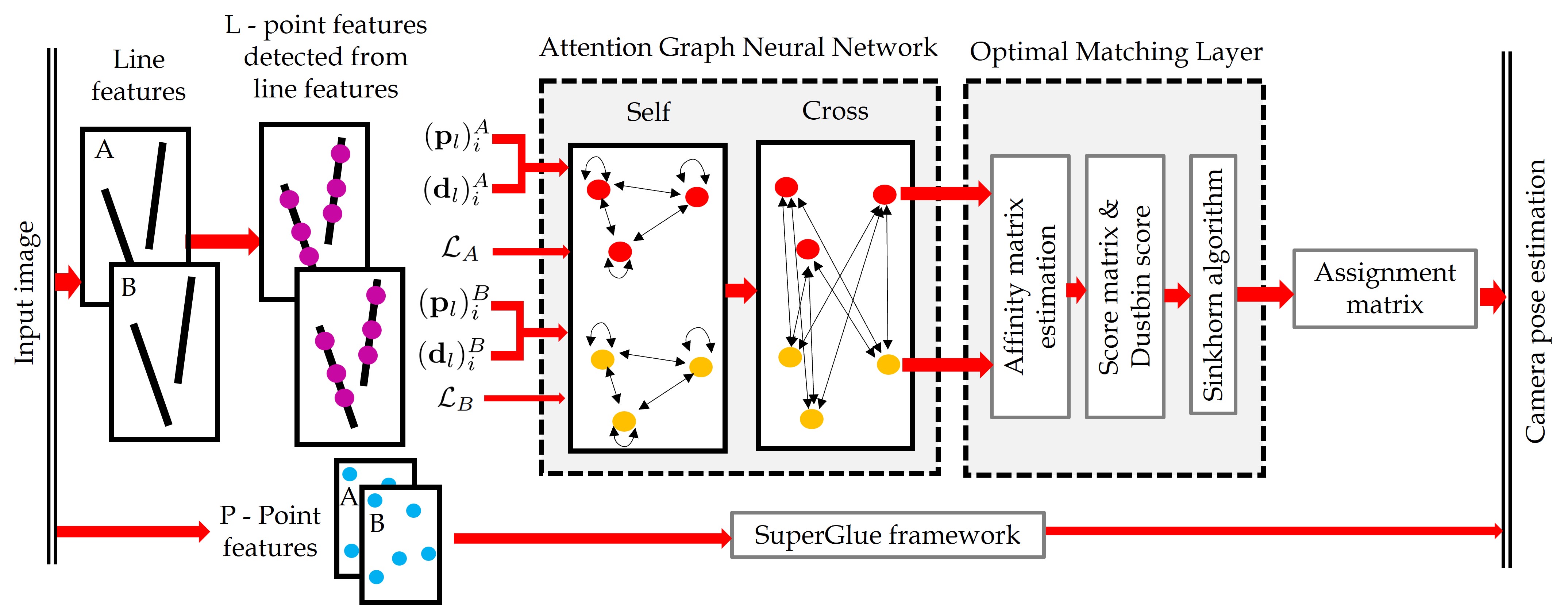}
  \caption{The proposed point and line feature matching architecture, consisting of point and line feature layers, an Attention Graph Neural Network layer for both points and lines, and an Optimal Matching layer that 
  generates an assignment matrix using the Sinkhorn algorithm.}
  \label{fig:framework}
\end{figure*}

We selected SOLD2 for line feature matching, 
since it has a similar CNN architecture 
to SuperPoint. 
To improve the performance of SOLD2 in low-light conditions and other adverse weather conditions, we fine-tuned the network on synthetic data generated using the CARLA driving simulator \cite{dosovitskiy2017carla}. Since SOLD2 performs well in ideal daytime conditions, we used the line features detected by SOLD2 
as ground truth and changed weather and lighting conditions for the same scenes in CARLA to 
generate multi-weather and lighting-augmented data. The SOLD2 algorithm also performs line matching by sampling lines and performing feature matching between the samples to aid in occlusion awareness. However, this results in incorrect matches in feature-poor scenarios. To overcome this, we introduced position constraints on the line features by sampling points 
along the lines, using SuperPoint to detect point features from these sets of sampled points,
and encoding the  point features with keypoint position awareness. This is explained further 
in the next sections. Both the point and line feature detection networks are designed to function effectively across various datasets 
without being overly reliant on specific training data. 
Their generalizability enables 
them to detect point and line features 
in diverse contexts and domains, making them versatile tools for a wide range of applications. Figure \ref{fig:framework} provides additional details of 
the framework. 

\subsection{Notation and definitions} \label{sec:NotationDef}

Consider a pair of 
images labeled $A$ and $B$. Each image $\alpha \in \{A, B\}$ has $L_\alpha$ line features, indexed by set $\mathcal{L}_\alpha \subset \mathbb{Z}_+$, 
and $P_\alpha$ point features, indexed by set $\mathcal{P}_\alpha \subset \mathbb{Z}_+$ and referred to as {\it P-point features}. 
The pixels comprising the line features are extracted using SOLD2 line feature extractor, 
and SuperPoint is used to detect $Q_\alpha$ point features from these pixels, indexed by set $\mathcal{Q}_\alpha \subset \mathbb{Z}_+$ and referred to as {\it L-point features}. 
Each point feature 
is associated with a 
position $\mathbf{p}$ and a visual descriptor vector $\mathbf{d}$. The feature 
position is defined as $\mathbf{p} = [u~v~c]^T$, where $u$ and $v$ are the pixel coordinates of the point and $c$ is the descriptor detection confidence. 
We will use the notation $(\mathbf{p}_p)_i^\alpha$ to indicate the position of P-point feature $i \in \mathcal{P}_\alpha$ in image $\alpha$ and  $(\mathbf{p}_l)_i^\alpha$ to indicate the position of L-point feature $i \in \mathcal{Q}_\alpha$ in image $\alpha$. We define the visual descriptors $(\mathbf{d}_p)_i^\alpha$, $(\mathbf{d}_l)_i^\alpha$ similarly. The subscript $x$ will refer to either $p$ or $l$.    

The representation $(\mathbf{y}_x)_i^\alpha$, $x \in \{p,l\}$, for each keypoint $i$ in image $\alpha$ is a high-dimensional vector that encodes 
the keypoint's position and visual descriptor. The keypoint position is embedded into $(\mathbf{y}_x)_i^\alpha$ 
 as follows using a multi-layer perceptron (MLP), similar to the SuperGlue architecture: 
\begin{equation}
    (\mathbf{y}_x)_i^\alpha = (\mathbf{d}_x)_i^\alpha + MLP_{encoder} (\mathbf{p}_x)_i^\alpha
\end{equation}

Our framework enforces the following constraints. (1) Any P-point feature 
from one image has exactly one match to a P-point feature in the other image; similarly for L-point features. 
(2) All P-point features and L-point features that are occluded or undetected will be unmatched. (3) 
A line feature $l_a \in \mathcal{L}_A$ in image $A$ is matched to a line feature $l_b \in \mathcal{L}_B$ in image $B$ if most of the L-point features on $l_a$ are matched to L-point features on $l_b$.

\subsection{Attention Graph Neural Network}


An Attention Graph Neural Network (GNN) forms the first layer of the architecture. The network encodes both the positions and visual descriptors of the keypoints, which ultimately improves the performance of the network over a conventional graph neural network.
The position constraints increase line-matching robustness and ensure that incorrect line matches do not occur in cases where images contain repetitive structures, such as windows in high-rise buildings. We developed separate GNNs 
for point and line feature matching, one with nodes defined as the P-point features and the other with nodes defined as the L-point features. Each GNN has a different set of losses and weights, since the networks 
 compute different estimates of the geometric and 
photometric cues. 

As in the SuperGlue architecture, aggression is achieved through both self- and cross-attention mechanisms. Given a feature that corresponds to a particular node in one image, self-attention aggregates features that correspond to adjacent nodes in the same image, and cross-attention aggregates similar features that correspond to nodes in another image.
The framework attends to individual point features' positions and their positions relative to
adjacent point features, as in SuperGlue. 
Let $(\mathbf{h}_x)_i^\alpha$, $x \in \{p,l\}$,  denote the matching descriptor for keypoint $i$ in image $\alpha$.  
The matching descriptors are defined as: 
\begin{equation}
    (\mathbf{h}_x)_i^\alpha = \mathbf{W}_x(\mathbf{y}_x)_i^\alpha + \mathbf{b}_x, ~~  
 \space x \in \space\{p, l\}, ~\alpha \in \{A, B\},
\end{equation}
where $\mathbf{W}_x$ is a weight matrix and $\mathbf{b}_x$ is a bias vector.

\subsection{Optimal Matching layer}

The Optimal Matching layer forms the second block of the framework, similar to SuperGlue. The input to this layer is the structural affinity between the two GNNs that have been encoded, defined in terms of affinity 
 matrices $\mathbf{S}_p \in \mathbb{R}^{P_A \times P_B}$ and $\mathbf{S}_l \in \mathbb{R}^{Q_A \times Q_B}$. The $(i,j)$-th entry of each matrix represents the affinity score
 between point feature $i$ in image $A$ and point feature $j$ in image $B$ and is defined as follows: 
\begin{equation} 
 (\mathbf{S}_x)_{i,j} = \text{exp}\left( \frac{(\mathbf{h}_x^{T})_i^A \space \mathbf{E}_x  \space (\mathbf{h}_x)_j^B}{\delta_x} \right), ~~x \in \{p,l\},
\end{equation} 
where $\mathbf{E}_x$ is a learnable weight matrix and $\delta_x$ is a tunable hyperparameter. 
The network is subject to the constraints described 
in  Section \ref{sec:NotationDef}. As in SuperGlue, the unmatched and occluded P-point and L-point features are assigned to a dustbin, which augments each affinity matrix with an additional row and column that are both filled with a single learnable parameter. 

We formulate the constrained optimization problem \eqref{eq:opt}-\eqref{eq:const} below to solve for the assignment matrices $\mathbf{P}_p \in \mathbb{R}^{P_A \times P_B}$ and $\mathbf{P}_l \in \mathbb{R}^{Q_A \times Q_B}$: 
\begin{equation} 
\text{max}\sum_{i=1}^{N+1} 
\sum_{j=1}^{M+1} 
(\mathbf{S}_{x})_{i,j} \space (\mathbf{P}_{x})_{i,j}, ~~x \in \{p,l\} 
\label{eq:opt}
\end{equation} 
\begin{equation}
\mathbf{P}_{x} \mathbf{1}_{N+1} = \mathbf{a} \space \text{~~and~~} \space {\mathbf{P}_{x}}^{T} \mathbf{1}_{M+1} = \mathbf{b},
    \label{eq:const}
\end{equation}
where $M = P_A$, $N= P_B$ for $x = p$; $M = Q_A$, $N = Q_B$ for $x=l$; and $\mathbf{a}$ and $\mathbf{b}$ are biases.
As in SuperGlue, this constitutes a differentiable optimal transport problem and can be solved using the Sinkhorn algorithm \cite{cuturi2013sinkhorn}, which is GPU-optimized. 
The algorithm is iterated until convergence.

Since all layers of the network are differentiable, we use the negative log-likelihood loss as the matching prediction loss. We backpropagate from ground truth matches to visual descriptors. Let $GT_p$ be the set of ground truth matches of P-point features, $\{(i,j)\} \subset \mathcal{P}_A \times \mathcal{P}_B$, and $GT_l$ be the set of ground truth matches of L-point features, $\{(i,j)\} \subset \mathcal{Q}_A \times \mathcal{Q}_B$. The sets $\mathcal{A}_p \subseteq \mathcal{P}_A$ and $\mathcal{B}_p \subseteq \mathcal{P}_B$ will denote the unmatched P-point features in both images, and similarly, $\mathcal{A}_l \subseteq \mathcal{Q}_A$ and $\mathcal{B}_l \subseteq \mathcal{Q}_B$ will denote the unmatched L-point features.
We define two losses, one for P-point features and one for L-point features:  
\begin{equation}
\begin{split}
Loss_x ~=~ -\sum_{(i,j) \space \in \space GT_x}^{} \log (\mathbf{P}_{x})_{i,j} - \sum_{i \space \in \space \mathcal{A}_x}^{} \log (\mathbf{P}_{x})_{i,M+1} \\
- \sum_{j \space \in \space \mathcal{B}_x}^{} \log (\mathbf{P}_{x})_{N+1,j}, ~~~~~x \in \{p,l\}, \nonumber
\end{split}
\label{eq:loss}
\end{equation}
where $M = P_A$, $N= P_B$ for $x = p$ and $M = Q_A$, $N = Q_B$ for $x=l$.

\subsection{Camera pose estimation}
 As a final step, we perform camera pose estimation by using the point and line matches from 2D-3D 
 point and line correspondences between successive frames. To obtain 3D points and lines, we perform stereo matching using the disparity map generated from the stereo camera images. Since this is a well-established topic and open source modules are readily available, a detailed description of camera pose estimation is outside the scope of this paper.

\begin{figure}[]
  \centering
  \includegraphics[width=\linewidth]{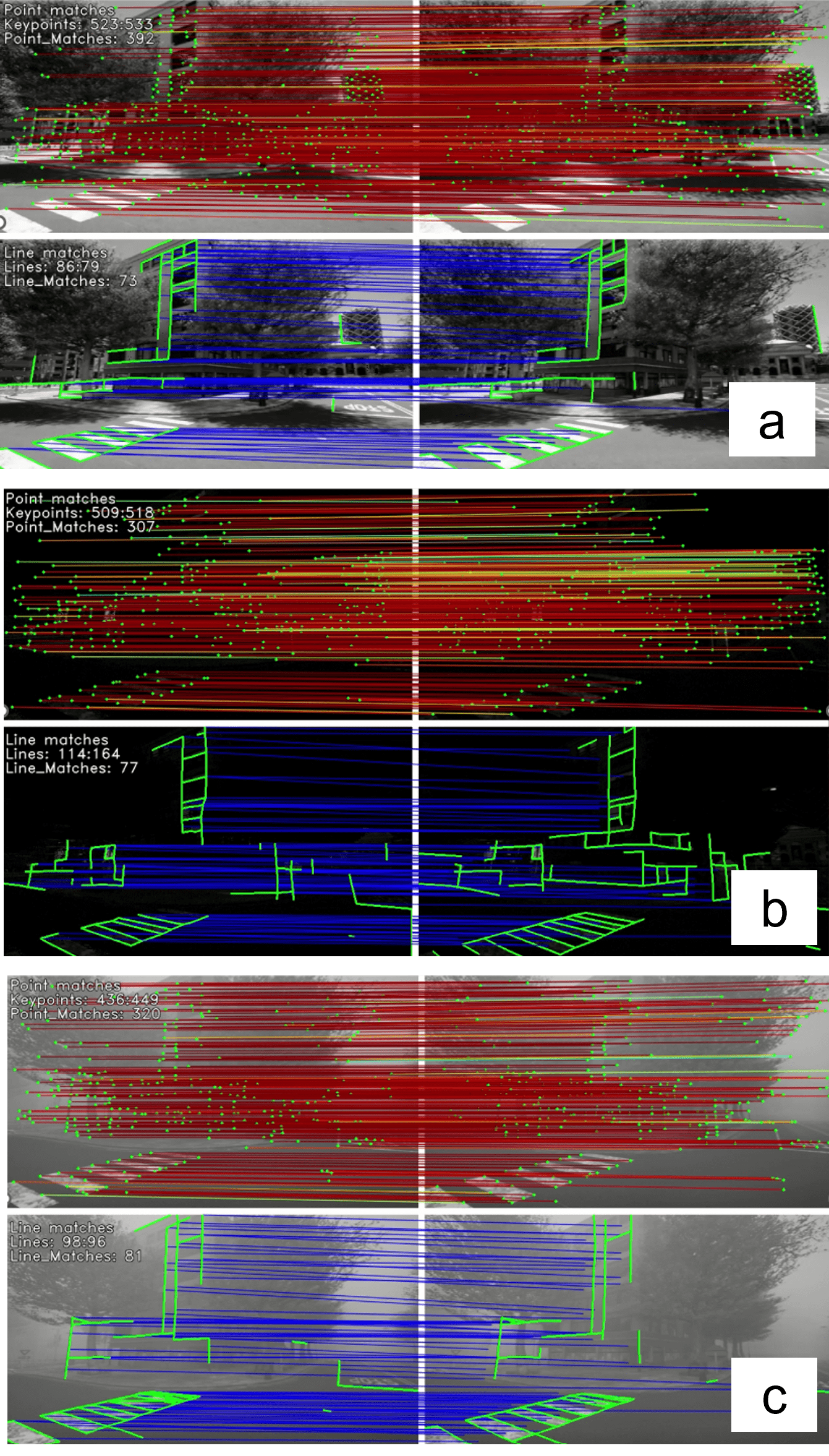}
  \caption{Every pair of rows starting from the top in this figure shows grayscale images of various synthetic scenarios from CARLA: 
  (a) clear sky scenario; (b) nighttime scenario; (c) foggy scenario. The first row shows point matches between frames $i$ and $i+1$,  and the second row shows line matches between frames $i$ and $i+1$. Point and line matches were generated by Method 2.} 
  \label{fig:3fig}
\end{figure}


\section{EXPERIMENTS} \label{sec:exp}






In this section, we present the results of experiments that compare the performance of our StereoVO framework 
to that of 
state-of-the-art algorithms for point and line feature matching. We compared \textbf{Method 1}, which combines the SuperGlue point-matching algorithm and SOLD2 line-matching algorithm, to \textbf{Method 2}, which combines the SuperGlue point-matching algorithm and our novel line-matching algorithm. We tested both methods 
on the following  datasets: 
\begin{itemize}
    \item Ford AV dataset~\cite{agarwal2020ford}, collected by autonomous vehicles, which consists of stereo camera images with accurate ground-truth trajectories obtained from LiDAR-based ICP. Our test data was drawn from Log 3 (Vegetation with clear sunny sky) and Log 4 (Residential area with clear sky). 
    \item Nighttime stereo camera images from the Oxford car dataset~\cite{RobotCarDatasetIJRR}, collected by an autonomous vehicle. Our test data consisted of images of residential areas with Visual Odometry (VO)-based ground truth.
    \item Synthetic stereo camera images with ground-truth trajectories from the urban environment in Town 10 of CARLA~\cite{dosovitskiy2017carla}. Our test data consisted of images of the same scenes under a variety of weather and lighting conditions, such as fog, nighttime, and glare. 
\end{itemize}  
Our StereoVO framework was run in real-time on an NVIDIA RTX 2080Ti GPU at around 7 FPS (142 ms). To ensure real-time operation, the framework requires
a minimum of 6 GB of GPU memory (VRAM).
Point and line matching results from Method 2 are shown in Fig. \ref{fig:teaser} and Fig. \ref{fig:3fig} for scenes from real and synthetic datasets, respectively, under various weather and lighting conditions.

\subsection{Comparison of estimated trajectories and pose error}

\begin{figure}[]
  \centering
  \includegraphics[width=\linewidth]{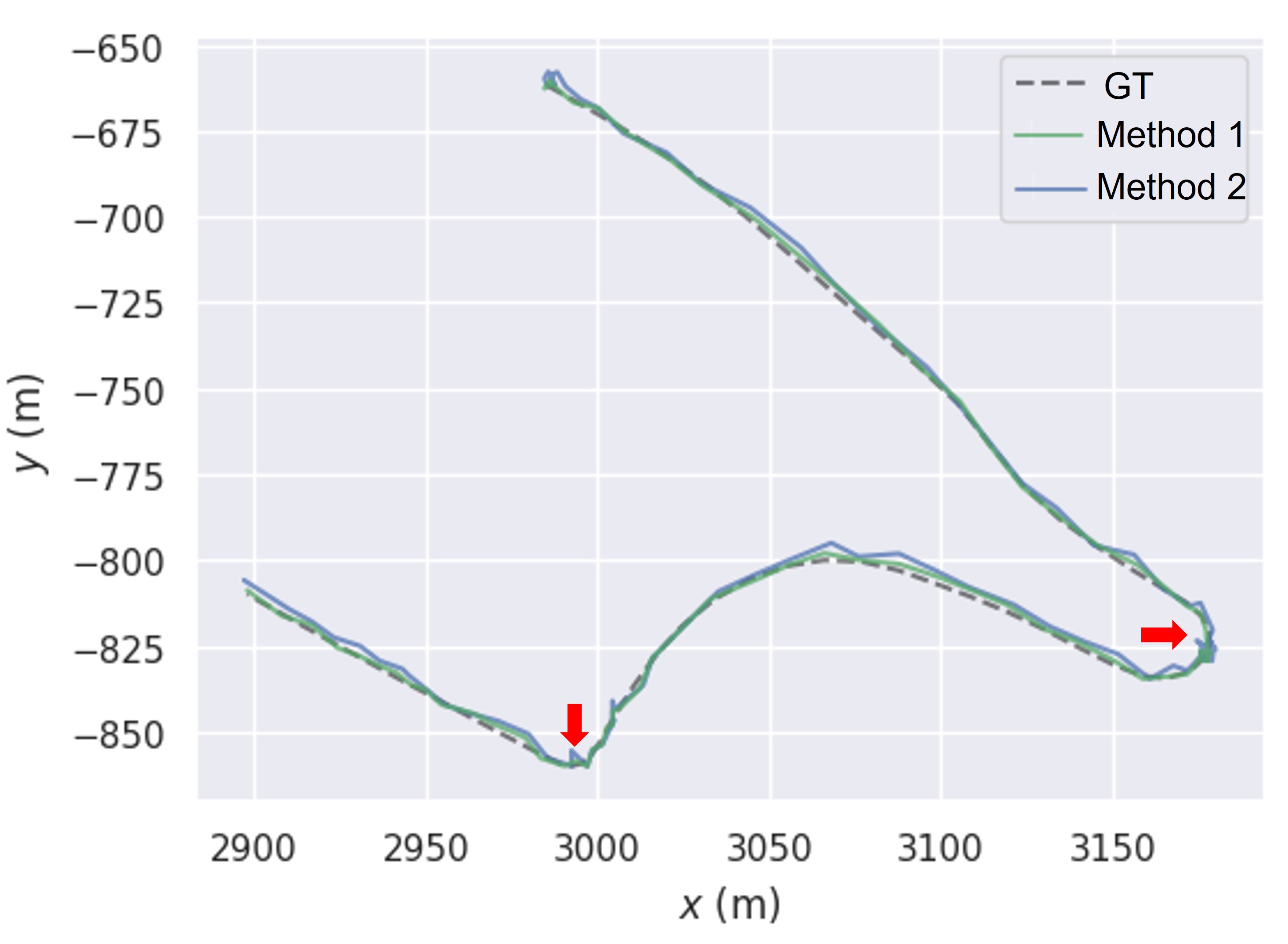}
  \caption{Comparison of Methods 1 and 2 on the Ford AV Log 4 dataset: ground truth (GT) trajectory and trajectories that were generated using camera pose estimates from each method.
  }
  \label{fig:traj}
\end{figure}

\begin{figure}[] 
  \centering
  \includegraphics[width=\linewidth]{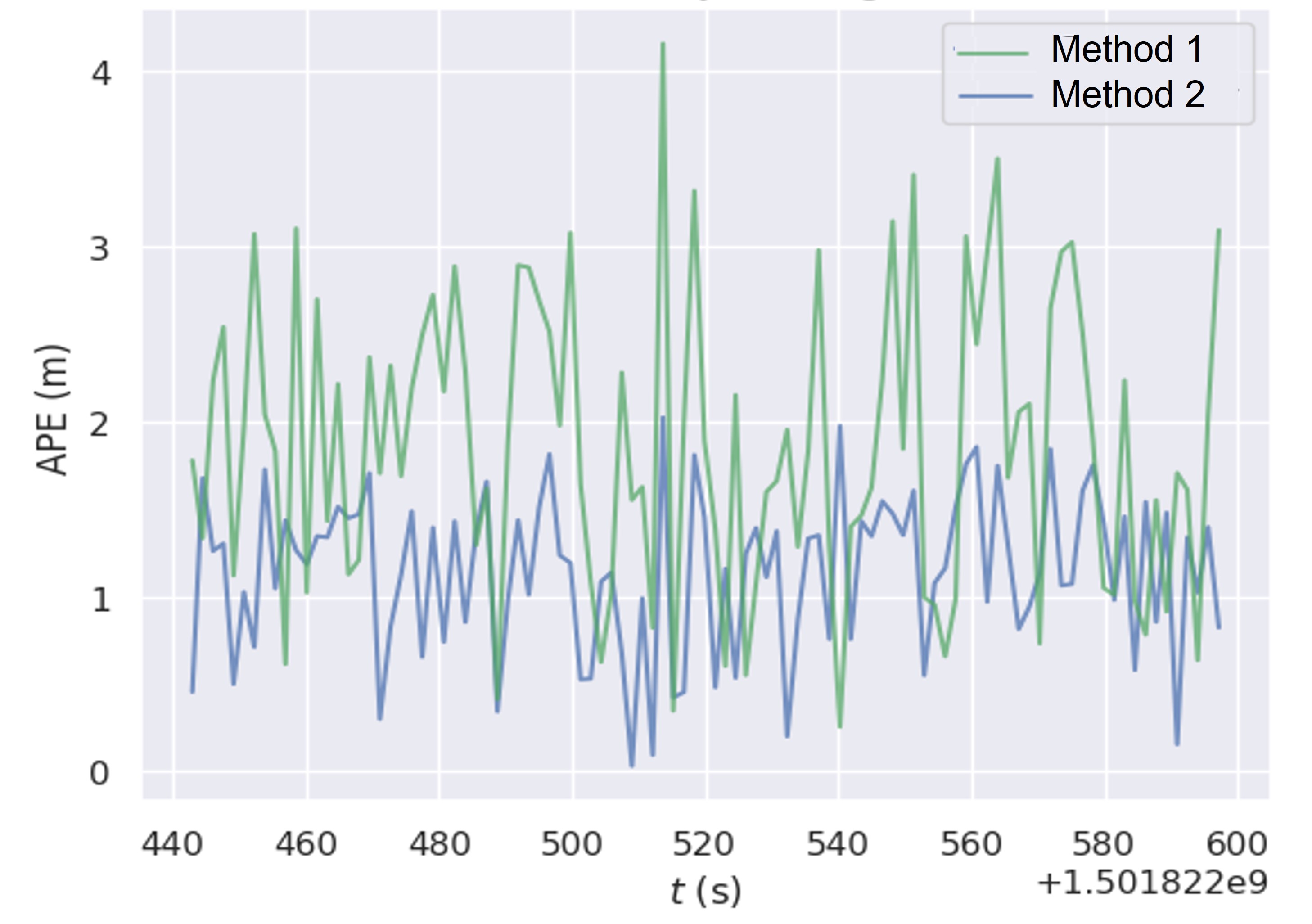}
  \caption{Comparison of Methods 1 and 2 on the Ford AV Log 4 dataset: Absolute Pose Error (APE) over a sample period of 160 s.
  The $x$-axis displays the 
  time recorded by the vehicle.}
  \label{fig:ape}
\end{figure}

We applied each method to estimate the vehicle camera poses from the Ford AV Log 4 dataset and generated vehicle trajectories from these pose estimates. The trajectories are plotted in Fig. \ref{fig:traj}, along with the ground truth (GT) trajectory. The figure shows that both methods yield trajectories that are close to the GT trajectory at all times. 
  Note that at coordinates 
 (2965, -660) and (3175, -825), indicated by arrows in the figure, the vehicle was at a complete stop, causing drift in the visual odometry. 
 This drift can be reduced by fusing measurements from other sensors, 
 such as GNSS and IMU, with the StereoVO estimates. 
 To quantify the deviation of the trajectory generated using each method from the GT trajectory, we computed the Absolute Pose Error (APE) over time
between each trajectory and the GT trajectory. Figure  \ref{fig:ape} compares the time series of the APE for both trajectories and shows that our method (Method 2) outperforms Method 1, in that it generally produced lower APE values over the sampled 160-s period. 

\begin{figure}[]
  \centering
  \includegraphics[width=\linewidth]{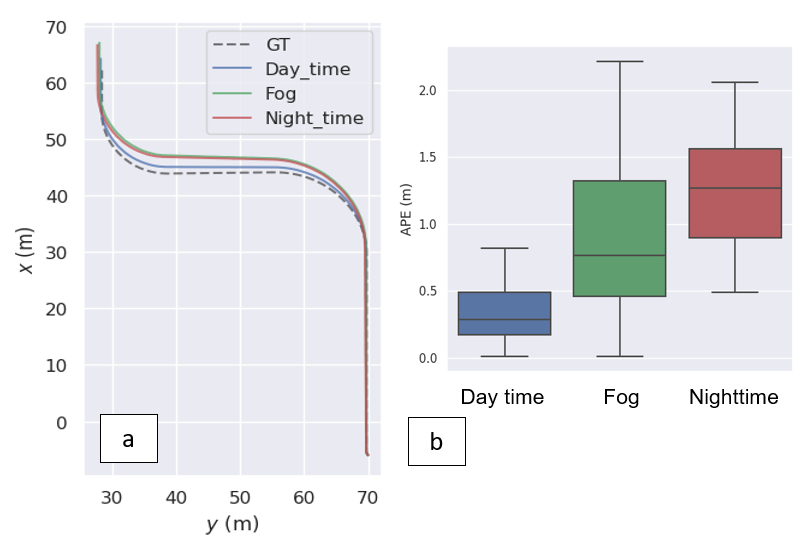}
  \caption{Evaluation of Method 2 on the CARLA synthetic dataset: (a) Ground truth (GT) trajectory and trajectories that were generated using camera pose estimates from images of the same scenes in CARLA under daytime, fog, and nighttime conditions.
  (b) Box plot of Absolute Pose Error (APE) over a sample of 200 frames for the same scenarios. 
  }
  \label{fig:traj2}
\end{figure}

We also evaluated the performance of Method 2 on synthetic data from CARLA for the following scenarios: (1) daytime with clear sky; (2) daytime with fog; and (3) nighttime with no street lights.  Figure \ref{fig:traj2}a plots the trajectory generated
over 200 frames for each scenario, along with the ground truth (GT) trajectory, and Figure \ref{fig:traj2}b shows the box plot of the absolute pose error (APE) for each scenario. Both figures indicate that the discrepancy between the estimated and GT trajectories is higher for the fog and nighttime scenarios than for the daytime scenario, as expected, but that all three estimated trajectories are relatively close to the GT trajectory.

Table \ref{table:RMSerror} reports the root-mean-square error (RMSE) 
between the estimated vehicle position and its ground truth position, obtained from simulated noise-free GNSS data, 
over 2000 frames in each of the three simulated scenarios in CARLA. The estimated positions were computed using Method 2 with only point features (detected by SuperPoint), only line features (obtained by our line-matching algorithm), or both point and line features.
For each scenario, the use of both point and line features yields a lower RMSE value than either point features or line features alone. Hence, 
the inclusion of line features in the StereoVO framework leads to improved performance, particularly in low-visibility and low-light 
conditions.

\begin{table}[]
\caption{RMSE between estimated and ground truth vehicle position for scenarios simulated in CARLA, using Method 2 with different combinations of point and line features
} \label{table:RMSerror}
\centering
\begin{tabular}{|c|c|c|c|} 
\hline
Scenario                    & Points & Lines &  RMSE (m) \\ \hline
\multirow{3}{*}{Daytime}  &   \checkmark     & -     &   0.2653            \\ \cline{2-4} 
                           & -      &    \checkmark    &    0.4312           \\ \cline{2-4} 
                           &   \checkmark      &    \checkmark    &   \textbf{ 0.1826 }          \\ \hline
\multirow{3}{*}{Fog}       &    \checkmark     & -     &    1.1563           \\ \cline{2-4} 
                           & -      &   \checkmark     &    1.1702           \\ \cline{2-4} 
                           &   \checkmark      &    \checkmark    &  \textbf{ 0.9865 }           \\ \hline
\multirow{3}{*}{Nighttime} &  \checkmark       & -     &   1.1840            \\ \cline{2-4} 
                           & -      &  \checkmark      &    1.1597           \\ \cline{2-4} 
                           &  \checkmark       &  \checkmark      &  \textbf{1.0168}            \\ \hline
\end{tabular}
\end{table}

\subsection{Comparison of number of feature detections and matches}

We also compared the number of point or line features that different algorithms detected and matched, along with the percentage of detected features that were matched, in 200 frames of the Ford AV, Oxford car, and CARLA datasets. Table \ref{table:matches_table} lists these quantities for point features that were detected by SuperPoint and matched by SuperGlue and line features that were detected by the SOLD2 line detector and matched by either the SOLD2 line-matching algorithm or ours. The table shows that our line-matching algorithm  recovers more line matches than the SOLD2 line-matching algorithm in each tested dataset. 

\begin{figure}[]
  \centering
  \includegraphics[width=\linewidth]{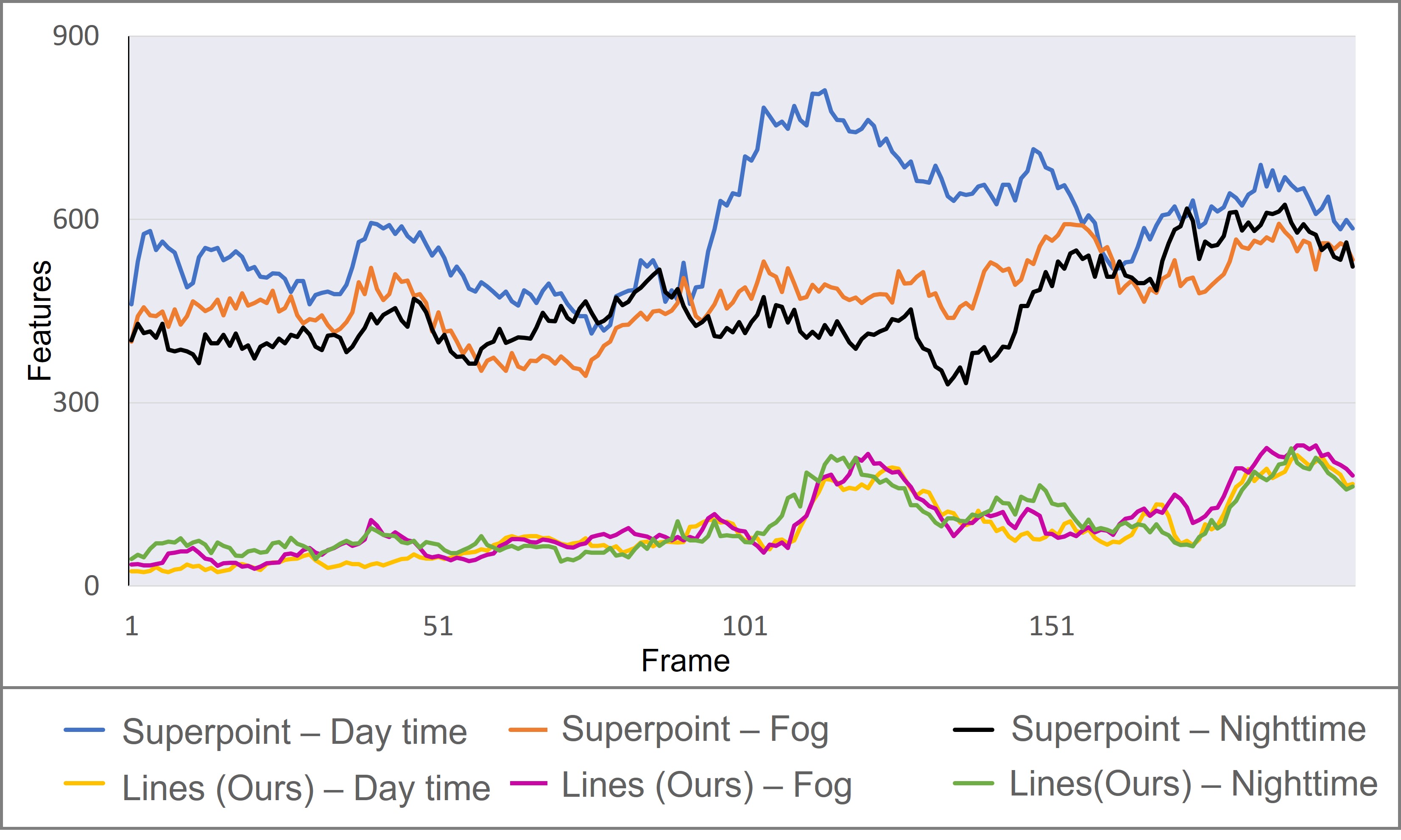}
  \caption{Number of point features detected by SuperPoint (top three plots) and number of line features obtained by our line-matching algorithm (bottom three plots) in each frame of three runs in CARLA under daytime, fog, and nighttime conditions.}
  \label{fig:graph}
\end{figure}

Figure \ref{fig:graph} plots the number of point features detected by SuperPoint and the number of line matches obtained by our algorithm in each frame of the daytime, fog, and nighttime scenarios simulated in CARLA. The figure shows that the number of line matches in each frame is not significantly affected by  the visibility conditions (clear or foggy) or light level (daytime or nighttime) in the scene. However, the numbers of point features detected in the fog and nighttime scenarios are consistently lower than the number detected in the daytime scenario, and are substantially lower in some frames. This is also reflected in the first row of Table \ref{table:matches_table}, which shows fewer total point detections in the CARLA fog and nighttime scenarios than in the daytime scenario.

These results indicate that our line-matching algorithm exhibits robust performance as a scene becomes more texture-poor due to adverse weather conditions and/or low illumination. In turn, this robustness in feature matching maintains the accuracy of the camera pose estimates under such conditions. 


\begin{table}[]
\caption{Number of feature detections (D) and matches (M) in real and synthetic datasets, using components of Methods 1 and 2
(L.F. = Line Features, L.M. = Line Matches)}
\label{table:matches_table}
\resizebox{\columnwidth}{!}{%
\begin{tabular}{|lc|ccc|ccc|}
\hline
\multicolumn{2}{|c|}{\multirow{2}{*}{Algorithms}}                                                               & \multicolumn{3}{c|}{Real datasets}                                                                                                                                                                                                                                  & \multicolumn{3}{c|}{\begin{tabular}[c]{@{}c@{}}Synthetic dataset \\ (CARLA)\end{tabular}}                                                                            \\ \cline{3-8} 
\multicolumn{2}{|c|}{}                                                                                       & \multicolumn{1}{c|}{\begin{tabular}[c]{@{}c@{}}Ford AV \\ dataset \\ (Log 3)\end{tabular}} & \multicolumn{1}{c|}{\begin{tabular}[c]{@{}c@{}}Ford AV \\ dataset \\ (Log 4)\end{tabular}} & \begin{tabular}[c]{@{}c@{}}Oxford \\ car \\ dataset\\ (Night)\end{tabular} & \multicolumn{1}{c|}{\begin{tabular}[c]{@{}c@{}}Day\\ time\end{tabular}} & \multicolumn{1}{c|}{Fog}            & \begin{tabular}[c]{@{}c@{}}Night\\ time\end{tabular} \\ \hline
\multicolumn{1}{|l|}{\multirow{3}{*}{\begin{tabular}[c]{@{}l@{}}SuperPoint +\\ SuperGlue\end{tabular}}} & D  & \multicolumn{1}{c|}{142563}                                                               & \multicolumn{1}{c|}{118642}                                                               & 22740                                                                      & \multicolumn{1}{c|}{132733}                                             & \multicolumn{1}{c|}{125774}          & 103457                                               \\ \cline{2-8} 
\multicolumn{1}{|l|}{}                                                                                  & M  & \multicolumn{1}{c|}{98364}                                                                & \multicolumn{1}{c|}{94237}                                                                & 17129                                                                      & \multicolumn{1}{c|}{116395}                                             & \multicolumn{1}{c|}{96972}          & 81304                                                \\ \cline{2-8} 
\multicolumn{1}{|l|}{}                                                                                  & \% & \multicolumn{1}{c|}{69.0}                                                                 & \multicolumn{1}{c|}{79.4}                                                                 & 75.3                                                                       & \multicolumn{1}{c|}{87.7}                                               & \multicolumn{1}{c|}{77.1}           & 78.6                                                 \\ \hline
\multicolumn{1}{|l|}{SOLD2 (L.F.)}                                                                      & D  & \multicolumn{1}{c|}{36634}                                                                & \multicolumn{1}{c|}{32145}                                                                & 8012                                                                       & \multicolumn{1}{c|}{17355}                                              & \multicolumn{1}{c|}{17840}          & 18272                                                \\ \hline
\multicolumn{1}{|l|}{\multirow{2}{*}{SOLD2 (L.M.)}}                                                     & M  & \multicolumn{1}{c|}{31089}                                                                & \multicolumn{1}{c|}{22061}                                                                & 3834                                                                       & \multicolumn{1}{c|}{12649}                                              & \multicolumn{1}{c|}{10426}          & 9046                                                 \\ \cline{2-8} 
\multicolumn{1}{|l|}{}                                                                                  & \% & \multicolumn{1}{c|}{84.9}                                                                 & \multicolumn{1}{c|}{68.6}                                                                 & 47.9                                                                       & \multicolumn{1}{c|}{72.9}                                               & \multicolumn{1}{c|}{58.4}           & 49.5                                                 \\ \hline
\multicolumn{1}{|l|}{\multirow{2}{*}{Our algorithm (L.M.)}}                                                      & M  & \multicolumn{1}{c|}{\textbf{34291}}                                                       & \multicolumn{1}{c|}{\textbf{28679}}                                                       & \textbf{6754}                                                              & \multicolumn{1}{c|}{\textbf{16861}}                                     & \multicolumn{1}{c|}{\textbf{16464}} & \textbf{17588}                                       \\ \cline{2-8} 
\multicolumn{1}{|l|}{}                                                                                  & \% & \multicolumn{1}{c|}{\textbf{93.6}}                                                        & \multicolumn{1}{c|}{\textbf{89.2}}                                                        & \textbf{84.3}                                                              & \multicolumn{1}{c|}{\textbf{97.1}}                                      & \multicolumn{1}{c|}{\textbf{92.3}}  & \textbf{96.2}                                        \\ \hline
\end{tabular}
}
\end{table}


\section{CONCLUSION}

We have presented a real-time stereo visual odometry framework aided by Attention Graph Neural Networks. This framework incorporates 
self-supervised learning-based point and line features and uses a novel line-matching technique that samples line features into 
point features with encoded position constraints. Using real datasets from autonomous vehicles and synthetic datasets from the CARLA driving simulator, we demonstrated that our framework produces robust line-matching in feature-poor scenes and scenes containing repetitive structures, e.g., Manhattan-world scenarios. In these tests, our framework outperformed state-of-the-art point and line feature matching algorithms in terms of the error between estimated and ground-truth vehicle poses, the percentage of detected line features that were matched, and the variability in  number of identified features with respect to visibility and lighting conditions.
One direction for future work is to incorporate planar features into the framework to improve its robustness. Moreover, developing a single end-to-end framework that performs temporal and stereo matching for both point and line features 
would be a promising step toward increasing the method's accuracy, robustness, and computation efficiency. 

\addtolength{\textheight}{-7cm}   

\bibliographystyle{./bibliography/IEEEtran}
\bibliography{./bibliography/IEEEabrv,./bibliography/Ref}

\begin{thebibliography}{10}
\providecommand{\url}[1]{#1}
\csname url@rmstyle\endcsname
\providecommand{\newblock}{\relax}
\providecommand{\bibinfo}[2]{#2}
\providecommand\BIBentrySTDinterwordspacing{\spaceskip=0pt\relax}
\providecommand\BIBentryALTinterwordstretchfactor{4}
\providecommand\BIBentryALTinterwordspacing{\spaceskip=\fontdimen2\font plus
\BIBentryALTinterwordstretchfactor\fontdimen3\font minus
  \fontdimen4\font\relax}
\providecommand\BIBforeignlanguage[2]{{%
\expandafter\ifx\csname l@#1\endcsname\relax
\typeout{** WARNING: IEEEtran.bst: No hyphenation pattern has been}%
\typeout{** loaded for the language `#1'. Using the pattern for}%
\typeout{** the default language instead.}%
\else
\language=\csname l@#1\endcsname
\fi
#2}}

\bibitem{agarwal2020ford}
S.~Agarwal, A.~Vora, G.~Pandey, W.~Williams, H.~Kourous, and J.~McBride,
  ``{Ford multi-AV} seasonal dataset,'' \emph{The International Journal of
  Robotics Research}, vol.~39, no.~12, pp. 1367--1376, 2020.

\bibitem{RobotCarDatasetIJRR}
W.~Maddern, G.~Pascoe, C.~Linegar, and P.~Newman, ``{1 Year, 1000km: The Oxford
  RobotCar Dataset},'' \emph{The International Journal of Robotics Research},
  vol.~36, no.~1, pp. 3--15, 2017.

\bibitem{c1}
D.~Lowe, ``Object recognition from local scale-invariant features,'' in
  \emph{Seventh IEEE International Conference on Computer Vision}, vol.~2,
  1999, pp. 1150--1157.

\bibitem{c2}
H.~Bay, T.~Tuytelaars, and L.~Van~Gool, ``{SURF: Speeded} up robust features,''
  in \emph{Computer Vision -- ECCV 2006}, A.~Leonardis, H.~Bischof, and
  A.~Pinz, Eds.\hskip 1em plus 0.5em minus 0.4em\relax Springer, 2006, pp.
  404--417.

\bibitem{c3}
E.~Rublee, V.~Rabaud, K.~Konolige, and G.~Bradski, ``{ORB: An efficient
  alternative to SIFT or SURF},'' in \emph{2011 International Conference on
  Computer Vision}, 2011, pp. 2564--2571.

\bibitem{c4}
D.~DeTone, T.~Malisiewicz, and A.~Rabinovich, ``{SuperPoint: Self-supervised}
  interest point detection and description,'' in \emph{IEEE Conference on
  Computer Vision and Pattern Recognition Workshops}, 2018, pp. 224--236.

\bibitem{c5}
K.~M. Yi, E.~Trulls, V.~Lepetit, and P.~Fua, ``{LIFT: Learned} invariant
  feature transform,'' in \emph{European Conference on Computer Vision}.\hskip
  1em plus 0.5em minus 0.4em\relax Springer, 2016, pp. 467--483.

\bibitem{Hausler2022}
S.~Hausler, M.~Xu, S.~Garg, P.~Chakravarty, S.~Shrivastava, A.~Vora, and
  M.~Milford, ``Improving worst case visual localization coverage via
  place-specific sub-selection in multi-camera systems,'' \emph{IEEE Robotics
  and Automation Letters}, vol.~7, no.~4, pp. 10\,112--10\,119, 2022.

\bibitem{voodarla2021}
M.~Voodarla, S.~Shrivastava, S.~Manglani, A.~Vora, S.~Agarwal, and
  P.~Chakravarty, ``{S-BEV: Semantic} birds-eye view representation for weather
  and lighting invariant 3-dof localization,'' \emph{arXiv preprint
  arXiv:2101.09569}, 2021.

\bibitem{pautrat2021sold2}
R.~Pautrat, J.-T. Lin, V.~Larsson, M.~R. Oswald, and M.~Pollefeys, ``{SOLD2:
  Self-supervised} occlusion-aware line description and detection,'' in
  \emph{IEEE/CVF Conference on Computer Vision and Pattern Recognition}, 2021,
  pp. 11\,368--11\,378.

\bibitem{abdellali2021l2d2}
H.~Abdellali, R.~Frohlich, V.~Vilagos, and Z.~Kato, ``{L2D2: Learnable} line
  detector and descriptor,'' in \emph{2021 International Conference on 3D
  Vision (3DV)}.\hskip 1em plus 0.5em minus 0.4em\relax IEEE, 2021, pp.
  442--452.

\bibitem{c7}
K.~L. Lim and T.~Br{\"a}unl, ``A review of visual odometry methods and its
  applications for autonomous driving,'' \emph{arXiv preprint
  arXiv:2009.09193}, 2020.

\bibitem{c8}
M.~He, C.~Zhu, Q.~Huang, B.~Ren, and J.~Liu, ``A review of monocular visual
  odometry,'' \emph{The Visual Computer}, vol.~36, 05 2020.

\bibitem{c9}
M.~O.~A. Aqel, M.~H. Marhaban, M.~I. Saripan, and N.~Ismail, ``Review of visual
  odometry: types, approaches, challenges, and applications,''
  \emph{SpringerPlus}, vol.~5, pp. 1--26, 2016.

\bibitem{c10}
D.~Bojani{\'c}, K.~Bartol, T.~Pribani{\'c}, T.~Petkovi{\'c}, Y.~D. Donoso, and
  J.~S. Mas, ``On the comparison of classic and deep keypoint detector and
  descriptor methods,'' in \emph{2019 11th International Symposium on Image and
  Signal Processing and Analysis (ISPA)}.\hskip 1em plus 0.5em minus
  0.4em\relax IEEE, 2019, pp. 64--69.

\bibitem{c11}
Y.~Ono, E.~Trulls, P.~Fua, and K.~M. Yi, ``{LF-Net: Learning} local features
  from images,'' \emph{Advances in Neural Information Processing Systems},
  vol.~31, 2018.

\bibitem{c6}
P.-E. Sarlin, D.~DeTone, T.~Malisiewicz, and A.~Rabinovich, ``{SuperGlue:
  Learning feature} matching with graph neural networks,'' in \emph{IEEE/CVF
  Conference on Computer Vision and Pattern Recognition}, 2020, pp. 4938--4947.

\bibitem{wang2009msld}
Z.~Wang, F.~Wu, and Z.~Hu, ``{MSLD: A} robust descriptor for line matching,''
  \emph{Pattern Recognition}, vol.~42, no.~5, pp. 941--953, 2009.

\bibitem{zhang2013efficient}
L.~Zhang and R.~Koch, ``An efficient and robust line segment matching approach
  based on {LBD} descriptor and pairwise geometric consistency,'' \emph{Journal
  of Visual Communication and Image Representation}, vol.~24, no.~7, pp.
  794--805, 2013.

\bibitem{vakhitov2019learnable}
A.~Vakhitov and V.~Lempitsky, ``Learnable line segment descriptor for visual
  {SLAM},'' \emph{IEEE Access}, vol.~7, pp. 39\,923--39\,934, 2019.

\bibitem{lange2019dld}
M.~Lange, F.~Schweinfurth, and A.~Schilling, ``{DLD: A} deep learning based
  line descriptor for line feature matching,'' in \emph{2019 IEEE/RSJ
  International Conference on Intelligent Robots and Systems (IROS)}.\hskip 1em
  plus 0.5em minus 0.4em\relax IEEE, 2019, pp. 5910--5915.

\bibitem{ma2020robust}
Q.~Ma, G.~Jiang, and D.~Lai, ``Robust line segments matching via graph
  convolution networks,'' \emph{arXiv preprint arXiv:2004.04993}, 2020.

\bibitem{zuo2017robust}
X.~Zuo, X.~Xie, Y.~Liu, and G.~Huang, ``Robust visual {SLAM} with point and
  line features,'' in \emph{2017 IEEE/RSJ International Conference on
  Intelligent Robots and Systems (IROS)}.\hskip 1em plus 0.5em minus
  0.4em\relax IEEE, 2017, pp. 1775--1782.

\bibitem{yang2019visual}
Y.~Yang, P.~Geneva, K.~Eckenhoff, and G.~Huang, ``Visual-inertial odometry with
  point and line features,'' in \emph{2019 IEEE/RSJ International Conference on
  Intelligent Robots and Systems (IROS)}.\hskip 1em plus 0.5em minus
  0.4em\relax IEEE, 2019, pp. 2447--2454.

\bibitem{xu2022eplf}
L.~Xu, H.~Yin, T.~Shi, D.~Jiang, and B.~Huang, ``{EPLF-VINS: Real}-time
  monocular visual-inertial {SLAM} with efficient point-line flow features,''
  \emph{IEEE Robotics and Automation Letters}, vol.~8, no.~2, pp. 752--759,
  2022.

\bibitem{xu2022leveraging}
B.~Xu, P.~Wang, Y.~He, Y.~Chen, Y.~Chen, and M.~Zhou, ``Leveraging structural
  information to improve point line visual-inertial odometry,'' \emph{IEEE
  Robotics and Automation Letters}, vol.~7, no.~2, pp. 3483--3490, 2022.

\bibitem{xie2021segformer}
E.~Xie, W.~Wang, Z.~Yu, A.~Anandkumar, J.~M. Alvarez, and P.~Luo, ``{SegFormer:
  Simple} and efficient design for semantic segmentation with transformers,''
  \emph{Advances in Neural Information Processing Systems}, vol.~34, pp.
  12\,077--12\,090, 2021.

\bibitem{lin2014microsoft}
T.-Y. Lin, M.~Maire, S.~Belongie, J.~Hays, P.~Perona, D.~Ramanan,
  P.~Doll{\'a}r, and C.~L. Zitnick, ``Microsoft {COCO: Common} objects in
  context,'' in \emph{Computer Vision -- ECCV 2014}.\hskip 1em plus 0.5em minus
  0.4em\relax Springer, 2014, pp. 740--755.

\bibitem{dosovitskiy2017carla}
A.~Dosovitskiy, G.~Ros, F.~Codevilla, A.~Lopez, and V.~Koltun, ``{CARLA: An}
  open urban driving simulator,'' in \emph{Conference on Robot Learning}.\hskip
  1em plus 0.5em minus 0.4em\relax PMLR, 2017, pp. 1--16.

\bibitem{cuturi2013sinkhorn}
M.~Cuturi, ``Sinkhorn distances: Lightspeed computation of optimal transport,''
  \emph{Advances in Neural Information Processing Systems}, vol.~26, 2013.

\end{thebibliography}
\end{document}